\documentclass[11pt]{article}
\usepackage[utf8]{inputenc}
\usepackage{graphicx}
\usepackage{amsmath}
\usepackage{hyperref}
\usepackage{amsfonts}
\usepackage{geometry}
\usepackage{natbib}
\usepackage{booktabs} 
\usepackage{subcaption} 
\usepackage{authblk} 
\title{Predicting Turn-Taking Outcomes in Multi-Party Conversation: Interpretable Modelling of Speech and Gaze Dynamics with Interpersonal Closeness}
\author[1,2]{Mark Dourado\thanks{Email: \href{mailto:mcldo@dtu.dk}{mcldo@dtu.dk}}}
\author[1]{Karim Haddad \thanks{Email: \href{khaddad@gn.com}{khaddad@gn.com}}}
\author[1]{Henrik Gert Hassager}
\author[3]{Stefania Serafin}
\affil[1]{Research \& Exploration, GN Store Nord, Ballerup, Denmark}
\affil[2]{Department of Architecture, Design and Media Technology, Aalborg University, Copenhagen, Denmark}
\affil[3]{Department of Engineering Technology and Didactics, Technical University of Denmark, Kongens Lyngby, Denmark}
\date{}

\begin{document}

\maketitle

\begin{abstract}
Smooth speaker transitions are fundamental to effective conversation and rely on an interlocutor's ability to predict when to enter the conversation. This ability depends on accurately interpreting and expressing the verbal and non-verbal cues that signal when a speaker wishes to take or relinquish the floor. The process becomes even more complex in noisy, natural, multi-party settings, with multiple interlocutors available. This study models how gaze and speech, together with perceived interpersonal closeness, signal conversational floor changes in free four-person dialogue. Using the GaMMA corpus, we trained logistic regression models using interpretable, behaviourally motivated features extracted before each turn-taking event to classify floor-transfer outcomes as gaps or overlaps. Predictors included gaze features such as transition motifs and behavioural contrasts, entropy, gaze-based addressee identity, and mutual gaze, alongside speech features derived from speaker loudness, as well as perceived interpersonal closeness (IOS) between speakers. Results show that gaze features capture predictive structure, and that combining them with loudness improves performance (ROC AUC = 0.76 ± 0.04). Loudness reflected speaker control, while gaze dispersion and addressing indexed listener readiness and competitive entry. Performance remained robust across noise conditions, indicating that gaze provides a complementary, noise-resilient cue to turn-taking dynamics.
\end{abstract}

\section{Introduction}

Human conversation follows an efficient system of \emph{turn-taking} that allows participants
to alternate speaking with minimal overlap or silence. The foundational model by Harvey Sacks \cite{Sacks1974} describes turns as composed of turn-constructional units (TCUs) that reach \emph{transition-relevance places} (TRPs), where speaker change becomes possible.  Evidence shows that across languages, turn transitions typically occur within a few hundred milliseconds \citep{Stivers2009}, far faster than the time required to plan a new utterance from scratch. This has led to the view that listeners continuously predict the timing of upcoming turn completions and begin preparing responses before the current turn ends \citep{LevinsonTorreira2015}.

Turn coordination relies on multiple linguistic and non-verbal cues that signal opportunities for a speaker change. Previous work has shown that prosodic features such as pitch declination and syllable lengthening \citep{Caspers2003,gravano2011turn}, syntactic completion and continuation \citep{ford1996,Sacks1974}, and pragmatic markers such as filled pauses or gaze behavior \citep{clark2002using,duncan1972some,ho2015speaking} all contribute to the timing of turn exchange. This timing can be expressed as the \emph{floor-transfer offset} (FTO), i.e. the temporal interval between one speaker’s offset and the next speaker’s onset, which can be quantified and defined as conversational gaps and overlaps \citep{heldner2010pauses}.

While the present study focuses on loudness \cite{stevens1955loudness} as the sole acoustic measure, these findings provide a broader context for how multimodal cues jointly shape the rhythm of conversation and can be modeled computationally for turn-taking prediction.

\subsection{Multimodal cues for predicting turns}

Face-to-face conversations integrate visual information alongside speech. 
Eye gaze plays a central role in regulating turns: speakers typically avert their gaze while holding the floor and direct it toward listeners as they near completion, thereby inviting uptake\citep{HollerKendrick2015}. 
In multiparty settings, gaze can also serve as a selection mechanism, with speakers looking toward an intended recipient at a TRP \cite{goodwin1981conversational}.
Conversely, gaze withdrawal and continued gesture often function as turn-holding devices, suppressing transition and discouraging listener self-selection until the speaker’s visual conduct signals availability \cite{HollerKendrick2015}.

Quantitative studies have confirmed these dynamics. 
In multi-party interaction, sequences of gaze shifts and head orientations predict both the next speaker and the timing of onset, often several hundred milliseconds before speech begins \citep{Ishii2013,Ishii2016}. 
Such findings indicate that predictive modelling of turn exchange should integrate visual information alongside acoustic and linguistic signals.

\subsection{Adaptation in noise and multimodal compensation}

Conversational coordination becomes even more complex in noisy environments where intelligibility is reduced. Speakers typically respond via the Lombard effect, increasing vocal intensity and modifying pitch or articulation to maintain audibility \citep{Summers1988,Garnier2010}. 
Previous work shows that conversational partners adapt their floor-transfer timing under noise, exhibiting slightly broader and right-shifted FTO distributions that indicate increased temporal variability and compensatory adjustments to maintain coordination under degraded acoustic conditions \citep{sorensen2021effects}.
These adaptations also extend to visual articulation and gestural behaviour, conceptually defining a multimodal Lombard effect \citep{Trujillo2021}. Under such conditions, both floor-yielding and floor-holding cues may be amplified as interlocutors strive to sustain coordination, while listeners compensate by orienting gaze more strongly toward speakers and relying on visual information \citep{Hadley2019, slomianka2025adaptions}.

\subsection{Toward modelling polyadic turn-taking}
Much empirical research has focused on dyadic- and triadic interactions, often under tightly controlled conditions. 
Polyadic conversation introduces greater complexity, with multiple potential next speakers, parallel dyadic behaviour (i.e. separate streams of conversation), and simply more competing bids for the floor. 
The \emph{GaMMA corpus} provides synchronised audio, gaze, and motion data from four-person groups recorded under both quiet and noisy conditions \citep{DouradoGaMMA2025,DouradoGaMMA_Dryad}. 
This resource enables the analysis of multimodal coordination in spontaneous, multi-party interaction. 
The present study uses these data to model how gaze and speech cues, together with perceived interpersonal closeness, predict turn-transition outcomes, specifically whether floor transfers occur as conversational gaps or overlaps.

\section{Floor-Transfer Offset and Conversational Event Detection}
\label{FTO}
\paragraph{Voice Activity Detection (VAD).}
Speech activity for each participant was estimated using a voice-activity detection (VAD) algorithm applied to individual microphone channels. 
This produced binary activity streams indicating when each participant was speaking, defining the temporal regions in which conversational events could occur.

\paragraph{Turns.}
Turns were derived from these VAD-based activity streams following the implementation by Sørensen~\cite{Sorensen2025}, which operationalises interpausal units (IPUs), gaps, and overlaps relative to floor changes. 
Here, a turn is defined as a period of continuous speech by a single speaker, bounded by transitions in floor ownership.

\paragraph{Floor-Transfer Offset (FTO).}
A central temporal descriptor of conversational timing is the \emph{floor-transfer offset} (FTO)—the interval between one speaker finishing and the next starting. 
FTO quantifies the coordination tightness between turns, capturing whether a transition occurs with silence (\emph{gap}) or temporal overlap (\emph{overlap}). 
It is defined only when the conversational floor changes from an outgoing speaker \(s\) to an incoming speaker \(t\) (\(t \neq s\)); backchannels and within-speaker overlaps are excluded. 
Let \(t^{\text{off}}_{s,i}\) denote the offset time of the outgoing speaker’s IPU at event \(i\), and \(t^{\text{on}}_{t,i}\) the onset time of the incoming speaker’s IPU. 
The signed FTO is

\[
\mathrm{FTO}_i = t^{\text{on}}_{t,i} - t^{\text{off}}_{s,i}.
\]

By convention, \(\mathrm{FTO}_i < 0\) indicates an \emph{overlap transfer} (incoming begins before outgoing ends) and \(\mathrm{FTO}_i \geq 0\) a \emph{gap transfer} (silence between turns), following the Communicative State Classification framework~\cite{Sorensen2025}.

\section{Gaze Data Representation and CSC Alignment}

\paragraph{Gaze representation.}
Using the GaMMA corpus \cite{DouradoGaMMA2025,DouradoGaMMA_Dryad}, each participant’s gaze direction was recorded continuously and represented as a 3D vector.  
Relative gaze angles were computed with respect to each other participants center of mass, per participant.
Additionally, each time sample was assigned to a discrete gaze state, which consisted of one of the other three co-participants (spanning P1–P4) or an \emph{Outside} category. Targets were considered active when the relative gaze angle fell within a \(\pm15^{\circ}\) threshold. 
A small hysteresis margin was applied to this threshold to prevent oscillations around boundary values. 
Short segments with near-constant angular velocity following a change of state between adjacent interlocutors (e.g. from a given speaker's left interlocutor, to the one situated on their right) were treated as transition artifacts from "grazing" the opposite speaker, and were thus pruned.
The resulting categorical gaze streams provide a stable, temporally aligned representation of visual attention for each participant.

\begin{figure}[!h]
    \centering
    \includegraphics[width=1\linewidth, trim = {0 0 0 1.5cm}, clip]{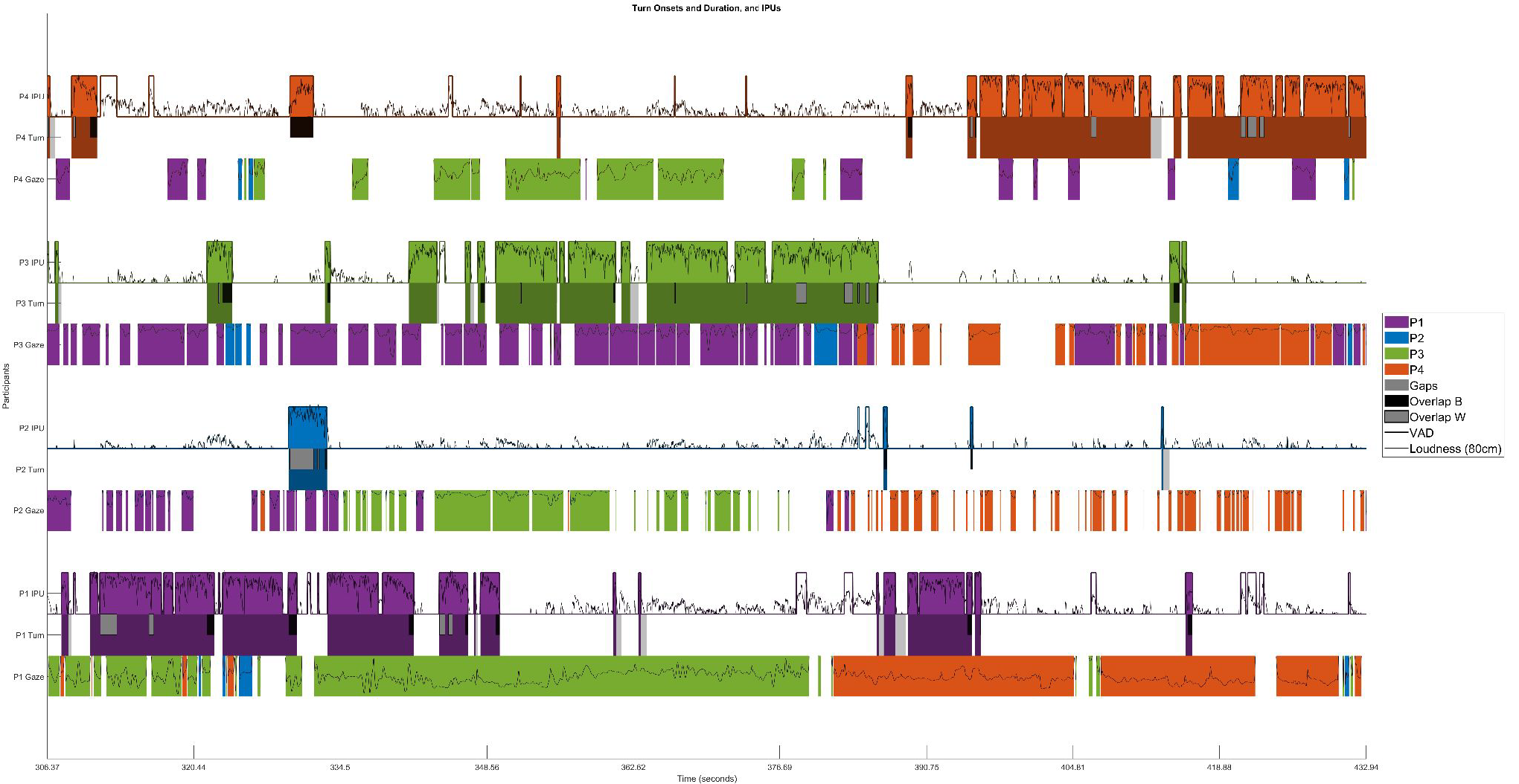}
    \caption{CSC representation of a 4-talker segment. For each participant Pn: Top row shows VAD per participant with IPU on/offsets and output level in Phon. Middle row shows labels, such as within-turn overlaps, gaps, and floor-times as well as detected floor changes (for FTO). Bottom: discretised gaze states showing who Pn is looking at, and relative gaze angle from centre-of-mass. The CSC view operationalises the inputs used for feature extraction and modelling.}
    \label{fig:CSC Plot}
\end{figure}

\paragraph{Communicative State Classification (CSC).}
Speech was labeled using the Communicative State Classification (CSC) framework \citep{Sorensen2025}, which consolidates per-participant close-mic speech activity,  segments and labels IPUs, FTOs, and floor changes.
Per-speaker loudness in Phon \cite{zwicker:iso} was synchronised to the CSC windows.
CSC operates as an alignment layer: it produces synchronised, machine-readable speech-state sequences that can be combined with the discretised gaze streams described above. 
Figure \ref{fig:CSC Plot} visualises the integrated representation, that serves as the basis for multimodal feature extraction in the subsequent analysis.

\newpage
\section*{Model Summary}

\subsection{Windowing and Alignment}
All features were computed within a single \emph{pre-turn} window aligned to each floor-transfer event. 
For every event \(i\), a 3\,s window was defined immediately preceding the turn-transition boundary, capturing multimodal activity leading up to the next speaker’s onset. 
Speech- and gaze-derived features were extracted within this window after temporal alignment to a common frame of reference, forming the input vector for subsequent modelling.

\subsection{Conceptual Model}
Each floor-transfer event \(i\) is represented by a model with the following feature families: 
\begin{equation}
\mathbf{x}_i =
\Big[
\mathrm{Phon}_{s},\;
\mathbf{C}^{(\mathrm{n\text{-}gram})},\;
\mathbf{M}^{(\mathrm{n\text{-}gram})},\;
\mathrm{SGE},\;
\mathrm{GTE},\;
\mathbf{IOS}_{s,t},\;
\mathbf{A}
\Big]_i
\label{eq:conceptual_model}
\end{equation}

\noindent\textbf{Feature family definitions.}
\begin{itemize}
  \item \(\mathrm{Phon}_{s}\): mean loudness (in Phon) for the outgoing speaker within the window and subwindows.
  \item \(\mathbf{C}^{(\mathrm{n\text{-}gram})}\): principal component scores (\textit{Contrasts}) from symbolic gaze n-gram sequences.
  \item \(\mathbf{M}^{(\mathrm{n\text{-}gram})}\): non-negative matrix factorisation (NMF) motif activations from symbolic gaze n-gram sequences.
  \item \(\mathrm{SGE}\), \(\mathrm{GTE}\): static and transition gaze entropies computed from gaze state transitions.
  \item $\mathrm{IOS}_{s,t}$:  Dyadic directional, perceived closeness scores for the active speaker and candidate target.
  \item \(\mathbf{A}_{s}\): categorical variables encoding addressee identity (seat index) and mutual gaze at end-of-IPU.
\end{itemize}

Two model variants were trained on identical folds and event sets:
\begin{enumerate}
    \item \textbf{Multimodal model}: The model as described in Eq.\ref{eq:conceptual_model}. 
    \item \textbf{Gaze-only model}: all gaze-derived features listed above, excluding \(\mathrm{Phon}_{s}\).
\end{enumerate}

\section{Methods}
\subsection*{AI Assistance Disclosure}
During manuscript preparation and analysis, AI-based tools (ChatGPT and GitHub Copilot) were used to assist with code debugging and generation, and to improve clarity, syntax, and condensation of the text in this article. All analyses, interpretations, and the final manuscript were designed, verified, and approved by the authors, who take full responsibility for its content.

\subsection*{Ethics Statement}
This study analyses data from the GaMMA corpus \cite{DouradoGaMMA_Dryad}. The original data collection was conducted with approval from the Aalborg University Research Ethics Committee (Case No. 2025-505-00609). All methods were performed in accordance with relevant guidelines and regulations, and all participants provided written informed consent, as detailed in the associated data descriptor publication \cite{DouradoGaMMA2025}. No new data were collected for the present study.

\subsection{Loudness Features}

Speech sound pressure levels were converted to perceived loudness in phons using the ISO~532-1 standard (Zwicker method) \cite{zwicker:iso}, applied to the calibrated close-microphone signals (denoised and cross–talk attenuated) provided in the GaMMA corpus \cite{DouradoGaMMA2025,DouradoGaMMA_Dryad}.\,
For each focal speaker \(s\) and floor-transfer event \(i\), framewise phon values were averaged within the window preceding the turn transition.
In addition to the mean loudness across the entire window, mean phon levels were computed for a series of 200 ms subwindows aligned to the window end. These short-term averages capture moment-to-moment variations in speaker intensity immediately prior to the transition.
To control for individual level differences, loudness values were normalised within speaker by z-scoring:

\[
\mathrm{Phon}_{i}^{(s)} = 
\frac{\mathrm{Phon}_{i}^{(s)} - \mu^{(s)}}{\sigma^{(s)}},
\]

where \(\mu^{(s)}\) and \(\sigma^{(s)}\) denote the mean and standard deviation of \(\mathrm{Phon}^{(s)}\) across all events for that speaker. The resulting \(\mathrm{Phon}_{i}^{(s)}\) values provide standardised measures of speaker loudness for use in the multimodal model.

\subsection{Social Feature: Interpersonal Closeness (IOS)}

Each participant rated their perceived interpersonal closeness to the other three interlocutors using the seven pictograms from the Inclusion of the Other in the Self (IOS) scale~\cite{aron1992IOS}. 
Let \(r^{(s \to t)}\) denote the raw IOS score that speaker \(s\) assigned to target \(t\) (\(t \neq s\)).
The IOS metric is static, dyadic, and directional; it does not vary across events or acoustic conditions.

\paragraph{Standardisation.}
To account for individual differences in scale use, ratings were standardised within each rater:
\[
z\mathrm{IOS}^{(s \to t)} = 
\frac{ r^{(s \to t)} - \mu^{(s)} }{ \sigma^{(s)} }, 
\qquad
\mu^{(s)} = \tfrac{1}{K_s}\!\!\sum_{u \neq s} r^{(s \to u)},
\]
where \(\sigma^{(s)}\) is the sample standard deviation across the \(K_s=3\) ratings provided by \(s\). 
If \(\sigma^{(s)}=0\), all standardised values for that participant were set to zero.

\paragraph{Event-level representation.}
For each floor-transfer event \(i\), let \(s\) be the outgoing speaker and \(t\) the candidate incoming speaker (either the next floor holder or the first overlap entrant). 
The standardised dyadic closeness scores in both directions are attached to the event as
\[
[\,z\mathrm{IOS}^{(s \to t)}_i,\; z\mathrm{IOS}^{(t \to s)}_i\,],
\]
which are static (time-invariant). Their sum and directional difference were additionally included as derived predictors.
\subsection{Symbolic Gaze Bigrams}

\paragraph{Rationale.}
To capture sequential structure in gaze behavior, we represented each participant’s gaze stream as a sequence of symbolic states and modeled their transitions using bigrams. 
Higher-order \(n\)-gram models (e.g., trigrams) provide sparse information in data due to the pre-turn window, as gaze often remains stable for periods exceeding the 3 seconds, resulting in sequences with no n-grams attached.
While bigrams are shorter, they ensured a higher yield and fewer empty sequences.

\begin{figure}[!h]
  \centering

  \begin{subfigure}[t]{0.72\textwidth}
    \centering
    \includegraphics[width=\textwidth, trim = {2cm 2cm 2cm 1.8cm}, clip]{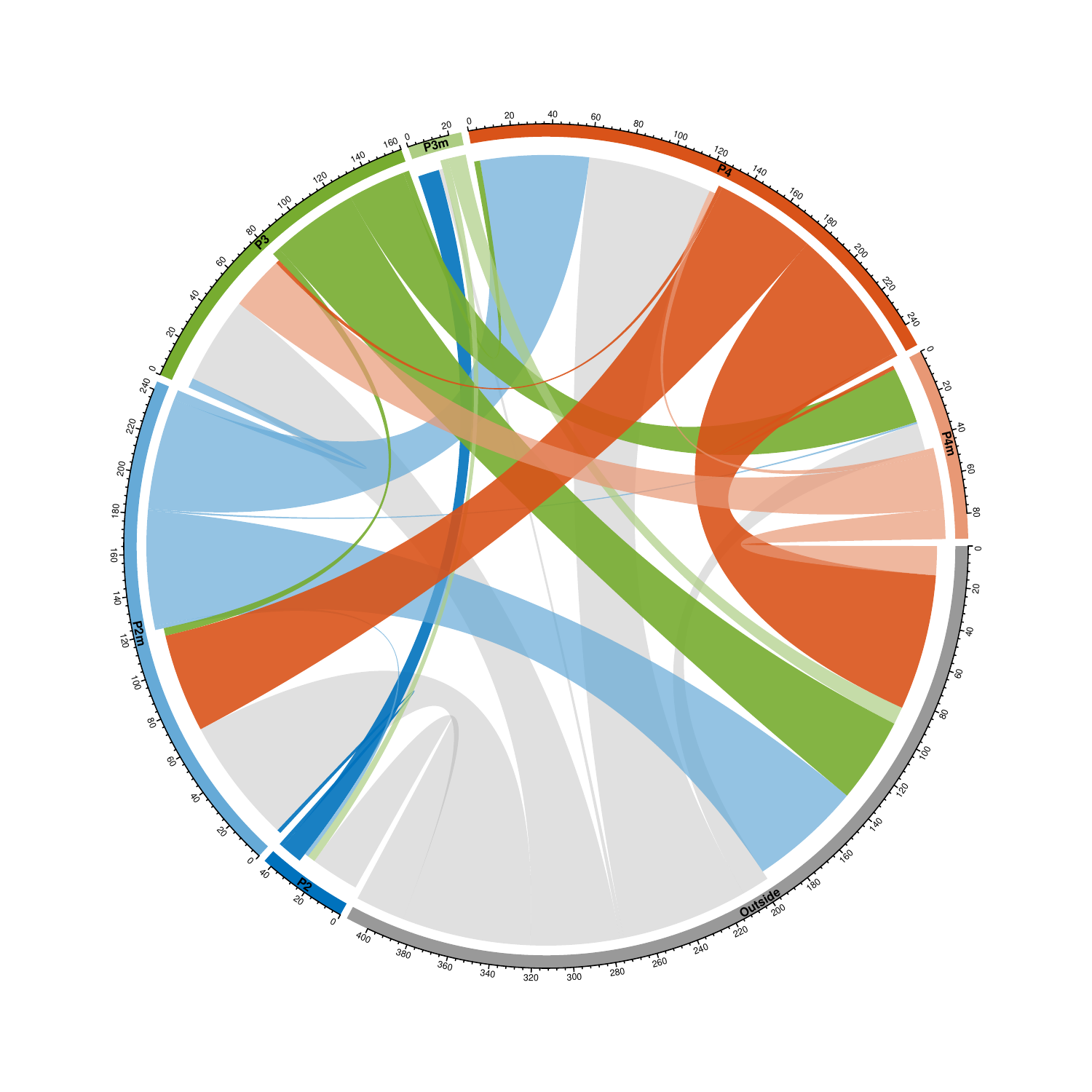}
    \subcaption{Participant 1}
    \label{fig:large_P1}
  \end{subfigure}
  \hfill
  \raisebox{82mm}{%
  \begin{minipage}[t]{0.25\textwidth}
    \begin{subfigure}[t]{\textwidth}
      \centering
      \includegraphics[width=\textwidth, trim = {0cm 3cm 0cm 2cm}]{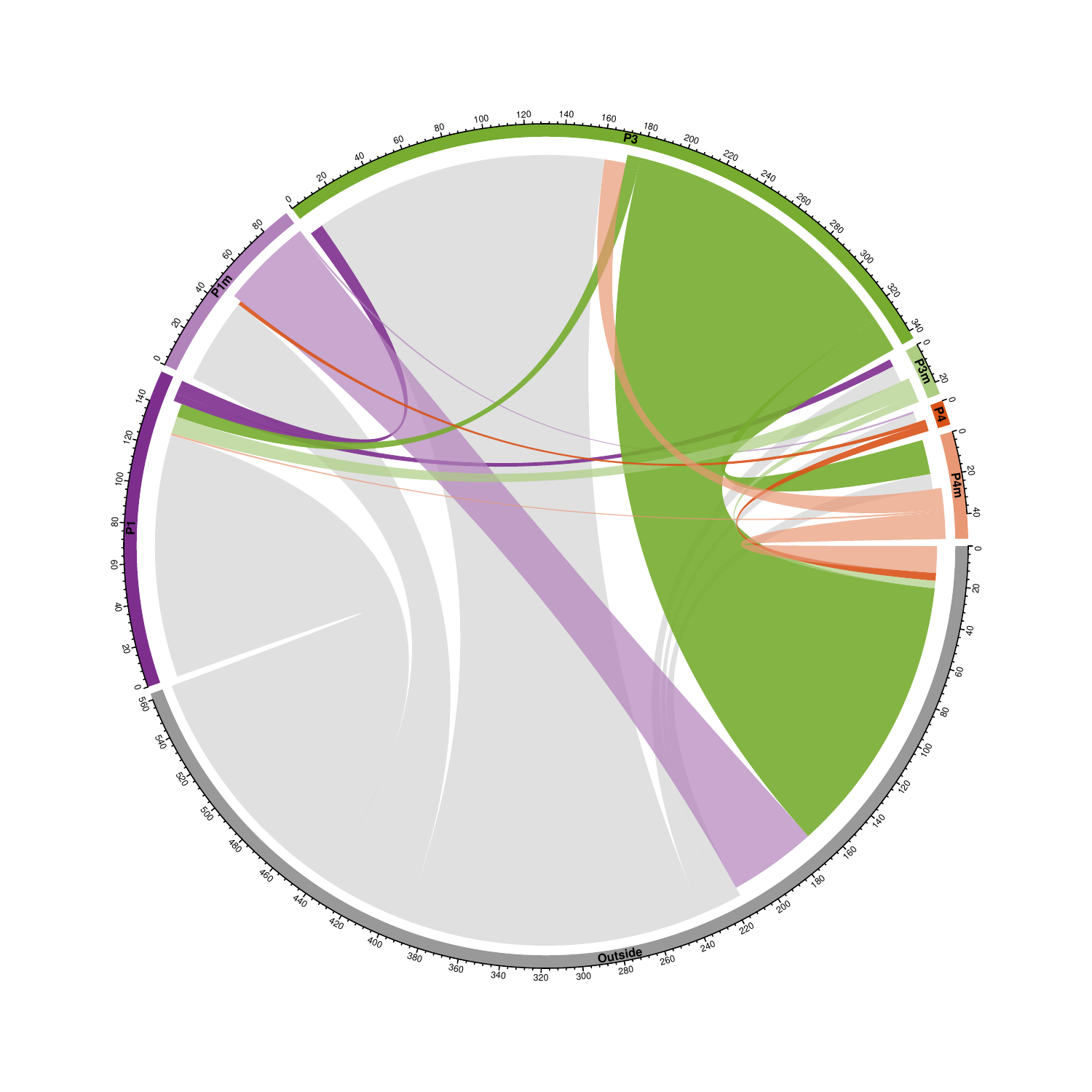}
      \subcaption{Participant 2}
      \label{fig:small_P2}
    \end{subfigure}

    \vspace{1mm}

    \begin{subfigure}[t]{\textwidth}
      \centering
      \includegraphics[width=\textwidth, trim = {0cm 3cm 0cm 2cm}]{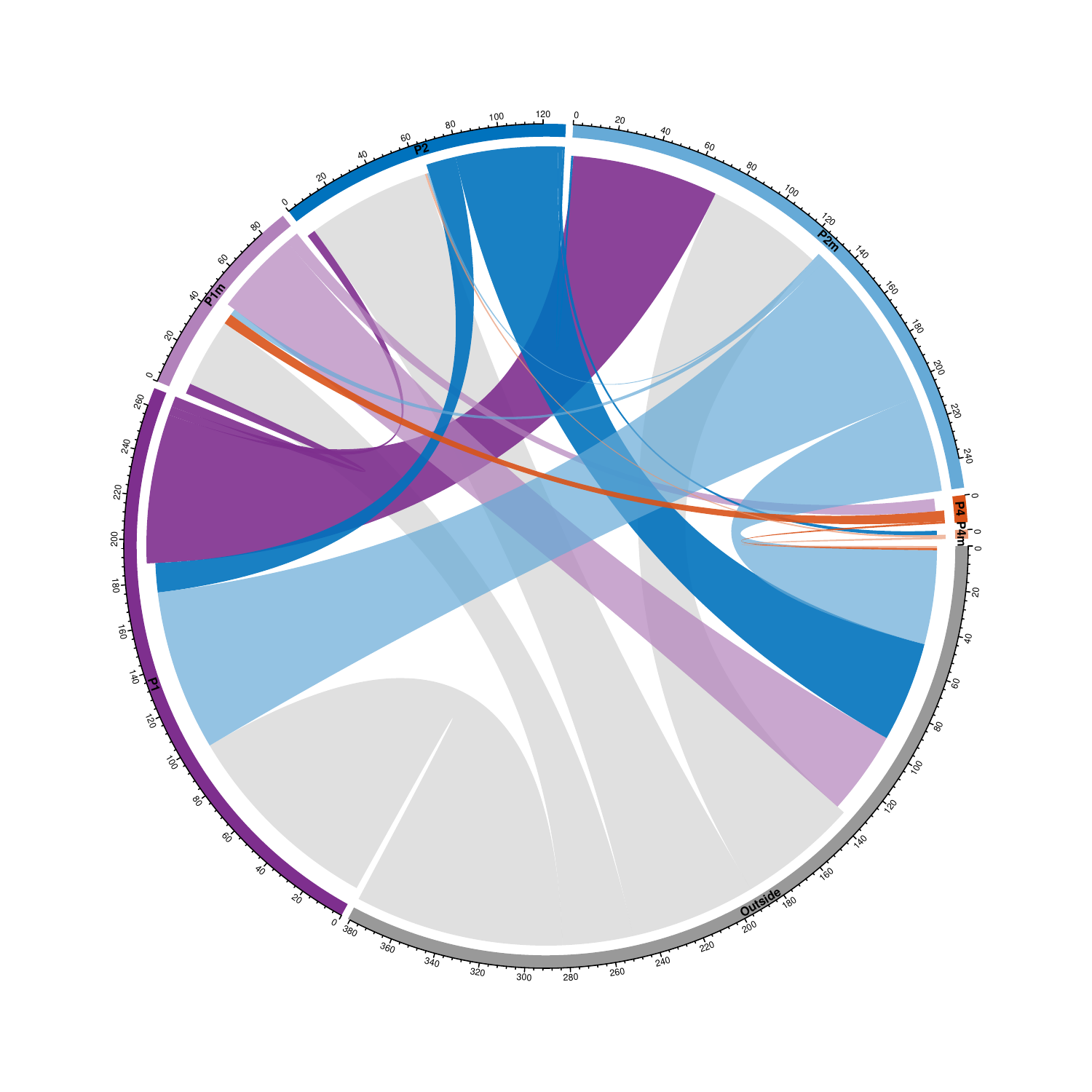}
      \subcaption{Participant 3}
      \label{fig:small_P3}
    \end{subfigure}

    \vspace{1mm}

    \begin{subfigure}[t]{\textwidth}
      \centering
      \includegraphics[width=\textwidth, trim = {0cm 3cm 0cm 2cm}]{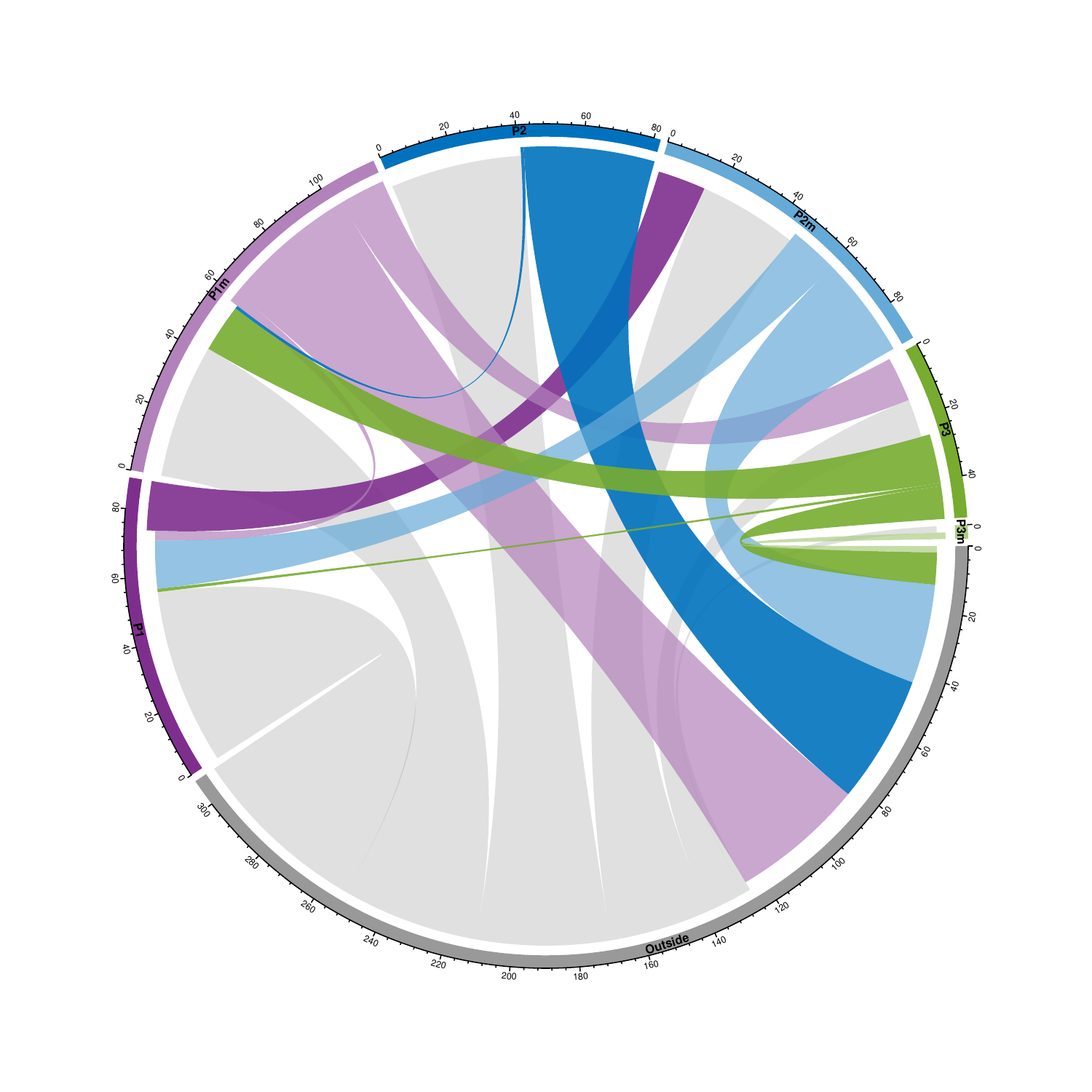}
      \subcaption{Participant 4}
      \label{fig:small_P4}
    \end{subfigure}
  \end{minipage}%
  }

  \caption{Representative bigram chord diagrams for one conversation: P1 (left) and P2–P4 (right). 
The diagrams illustrate the seven-state gaze space and transition frequencies between gaze targets. 
Chord thickness indicates transition frequency, while the relative gap size at the ring edge distinguishes the origin and destination of each bigram.}

  \label{fig:chord_left_large_right_strip}
\end{figure}

\paragraph{State space and transition representation.}
For each participant \(P_s \in \{P_1,P_2,P_3,P_4\}\), gaze behavior was represented in a seven-state space:

\begin{equation}
\mathcal{S}_s = \{\texttt{Outside},\, P_{t_a},\, P_{t_b},\, P_{t_c},\, P_{t_a}^m,\, P_{t_b}^m,\, P_{t_c}^m\},
\label{eq:addressing_states}
\end{equation}

where \(P_{t_j}\) denotes gaze directed toward interlocutor \(t_j\) and \(P_{t_j}^m\) indicates mutual gaze (reciprocal fixation).
Successive identical states were collapsed (run-length smoothing) before extracting adjacent pairs \((x^{(s)}[n], x^{(s)}[n+1])\).
Transitions were counted in a \(7\times7\) matrix \(B^{(s)}\), with entries normalised to yield relative frequencies:
\[
p^{(s)}(u,v) = 
\frac{B^{(s)}(u,v)}{\sum_{u',v'} B^{(s)}(u',v')}.
\]
The flattened transition matrix forms a 49-dimensional feature vector \(\mathbf{b}^{(s)} \in \mathbb{R}^{49}\) for participant \(s\), used as input to subsequent dimensionality-reduction steps. The transition matrices and state-space are visualised in figure \ref{fig:chord_left_large_right_strip}.

\newpage
\subsection{Dimensionality Reduction: PCA and NMF}
Due to their nature, many bigram transition frequencies are correlated, e.g: $P_s \!\to\! P_t^m$ often covaries with $P_t \!\to\! P_s^m$ through reciprocal gaze, resulting in a highly collinear feature space. 
To obtain compact and interpretable representations of gaze transition structure, we applied Principal Component Analysis (PCA) and Non-Negative Matrix Factorisation (NMF) within each cross-validation fold. 
All decompositions were performed fold-wise to avoid information leakage: training data were used to fit the basis, and test data were projected using parameters learned from the training subset only.

\paragraph{PCA.}
PCA was applied to symbolic gaze bigram matrices both at the participant level and at the group-aggregated level. 
Training matrices were centered by their fold-specific mean and decomposed into component loadings and scores. 
The first $k=5$ components were retained (> 90\% of variance), providing low-dimensional representations of recurrent transition \textit{contrasts} such as mutual gaze exchange or disengagement toward \texttt{Outside}.

\paragraph{NMF.}
NMF was applied to the global bigram matrices using multiplicative-update optimisation (rank $r=4$–$5$, five replicates). 
Training data $\mathbf{X}_{\mathrm{tr}}$ were factorised as $\mathbf{X}_{\mathrm{tr}} \approx \mathbf{W}_{\mathrm{tr}}\mathbf{H}$, and test data were projected via non-negative least squares using the fixed basis $\mathbf{H}$. 
This yielded \textit{motif} activations $\mathbf{W}$ describing interpretable gaze transition structures shared across participants.

\paragraph{Cross-validation integrity.}
Each fold (see \ref{Model Training and Evaluation}) produced an independent PCA/NMF basis derived exclusively from its training data. 
Test projections used these fold-specific parameters, ensuring complete separation between training and evaluation data. 

\paragraph{Interpretation.}
PCA and NMF provide complementary representations of gaze transition structure: PCA yields orthogonal, variance-based components that reduce collinearity, whereas NMF produces additive, nonnegative motifs that capture recurrent transition patterns. 
Features from both decompositions were included as alternative gaze-derived predictors in the regression models to evaluate efficiency versus interpretability.
\subsection{Static and Transition Gaze Entropy}

\paragraph{Motivation.}
Gaze behavior in spontaneous conversation is highly variable: moment-to-moment shifts can reflect attention, engagement, or planning. 
Entropy measures provide compact summaries of this complexity. 
We computed two complementary variants: \textit{Static Gaze Entropy} (SGE) and \emph{Gaze Transition Entropy} (GTE). They respectively quantify the dispersion of gaze targets and the unpredictability of gaze transitions \citep{krejtz2015entropy, shiferaw2019gaze}. 
Together, they capture both how broadly attention is distributed and how structured gaze dynamics are during turn exchange.

\paragraph{Computation.}
For each participant \(s\) and floor-transfer event \(i\), gaze samples within the pre-turn window were discretised into the seven-state space \(\mathcal{S}_s\) (Eq.~\ref{eq:addressing_states}).  
Static Gaze Entropy was computed as Shannon entropy over the empirical state distribution:
\[
\mathrm{SGE}^{(s)}_{i} = -\sum_{u \in \mathcal{S}_s} p(u)\,\log_{2} p(u),
\]
where \(p(u)\) is the proportion of samples assigned to state \(u\).
Gaze Transition Entropy was computed analogously from the bigram transition probabilities \(p(u,v)\) derived from the run-length-collapsed gaze sequence:
\[
\mathrm{GTE}^{(s)}_{i} = -\sum_{u,v \in \mathcal{S}_s} p(u,v)\,\log_{2} p(u,v).
\]
Lower SGE values indicate sustained fixation on few targets; higher values indicate dispersed attention. 
Lower GTE reflects repetitive or highly predictable gaze routines, whereas higher GTE denotes more variable and less predictable transitions.

\paragraph{Usage.}
Both \(\mathrm{SGE}^{(s)}_{i}\) and \(\mathrm{GTE}^{(s)}_{i}\) were computed for each participant and as group-aggregated measures within the window.

\subsection{Gaze Addressing Features}

\paragraph{Addressee estimation.}
For each turn-ending event, the speaker’s gaze direction was averaged over the final 600\,ms preceding floor transfer. 
If the mean gaze angle toward a target interlocutor fell within a 15$^\circ$ cone, that person was designated as the addressee; otherwise, the event was labeled as having no addressee. 
When mutual gaze occurred within the same window, the addressee label was shifted to a corresponding mutual-gaze category, yielding nine discrete states 
(same as in Eq.~\ref{eq:addressing_states}). 
Note that each participant’s local gaze space \(\mathcal{S}_s\) in this case is $n=9$, since it formally includes a self-state. This state is never realised in practice, as self-directed gaze is undefined.

\paragraph{Derived features.}
From these states we derived three complementary predictors:
(i) a categorical addressee identity (\textit{seat index}),
(ii) a binary mutual-gaze flag, and
(iii) a continuous addressed-weight feature quantifying the relative outcome likelihood associated with each addressee, computed fold-wise to prevent information leakage. 
The continuous feature was regularised across folds to stabilise estimates and centered to isolate addressing effects beyond global outcome rates. 
Together, these measures capture both categorical and graded aspects of visual addressing, linking gaze behavior to subsequent floor-transfer outcomes.

The addressed-weight feature (iii) was defined as the centered empirical outcome likelihood associated with the addressee:
\[
\mathrm{Addressed}^{(\mathrm{Weighted})}_i = \hat{p}_{a_i} - \bar{p},
\]
where $\hat{p}_{a_i}$ is the fold-specific probability of a gap outcome when addressing $a_i$, and $\bar{p}$ is the mean gap rate in the same fold.

\paragraph{Summary.}
The gaze-derived features collectively provide complementary perspectives on how visual attention is organised around floor-transfer events. 
The symbolic state space with mutual gaze (Eq.~\ref{eq:addressing_states}) forms the foundation from which we derive bigram matrices, dimensionality-reduced components (PCA and NMF), entropy measures (SGE and GTE), and categorical addressee identities. 
Bigram-based reductions capture recurrent motifs of gaze transitions, entropy quantifies the dispersion and unpredictability of gaze allocation, and addressee coding identifies the gaze target at turn endings. 
Together, these representations yield compact yet interpretable descriptions of gaze behavior, which are subsequently integrated with loudness and interpersonal features (IOS) in the multimodal regression models.


\section{Model Representation and Training}

\subsection{Outcome Variable}
The dependent variable for all classification models is the binary \emph{floor-transfer outcome}
$y_i^{\mathrm{FTO}}$, derived from the signed floor-transfer offset defined in
Section~\ref{FTO}.
Events with $\mathrm{FTO}_i \ge 0$ are labeled \textsc{Gap} ($y_i^{\mathrm{FTO}}=1$),
and events with $\mathrm{FTO}_i < 0$ are labeled \textsc{Overlap} ($y_i^{\mathrm{FTO}}=0$).
Labels are assigned only for genuine floor changes; backchannels and within-speaker overlaps are excluded.

\subsection{Feature Representation}
For each floor-transfer event $i$, the multimodal feature vector $\mathbf{x}_i$
concatenates all derived predictors: phonetic intensity (\textit{Phon}),
gaze-based features (SGE, GTE, PCA, NMF),
interpersonal closeness (\textit{IOS}),
and addressing-related variables (categorical addressee, mutual gaze flag, and addressed weight).
All features were computed independently within folds to ensure cross-validation integrity.
PCA and NMF bases were fitted on the training subsets only, and test events were projected using the fold-specific parameters to prevent information leakage.
Each predictor was z-scored using the mean and standard deviation of the corresponding training data before model fitting.

\subsection{Model Training and Evaluation}
\label{Model Training and Evaluation}
Each model was a regularised logistic regression with an elastic-net penalty
($\alpha = 0.5$), combining $L_1$ and $L_2$ regularisation.
The regularisation strength ($\lambda$) was tuned by inner five-fold cross-validation
within each outer fold.
Outer validation used both leave-one-group-out (LOGO) and leave-one-condition-out (LOCO) schemes,
yielding 15 outer folds in total.
Performance on held-out folds was evaluated using
ROC AUC, PR AUC, F1-score, Precision, Recall, and Accuracy.

\section{Results}

\subsection{Overall Predictive Performance} \label{Overall Predictive Performance}
\begin{figure}[!h]
    \centering
    \includegraphics[width=1\linewidth, trim = {0 0 0 0cm}, clip]{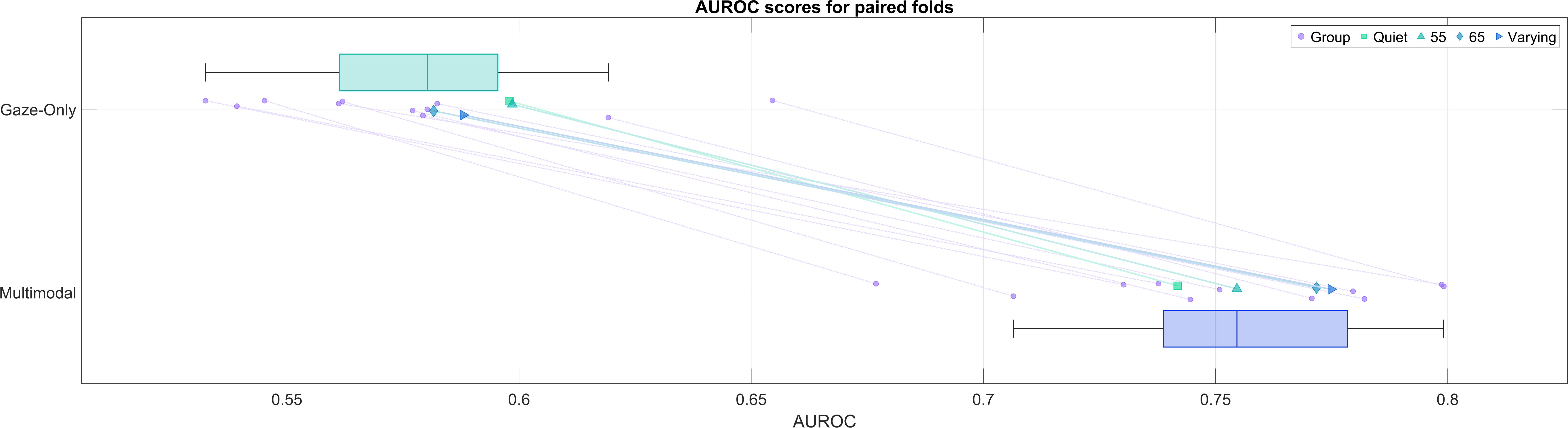}
    \caption{Paired fold-level AUC (AUROC) for gaze-only and multimodal models. 
    Each line represents one cross-validation fold (LOGO/LOCO). 
    The multimodal model consistently outperformed the gaze-only model ($\Delta$AUC mean~=~0.18; $p<0.001$).}
    \label{fig:AUC_ModelPerf_Gaze_Phon}
\end{figure}

We evaluated how well each model discriminated between \textit{gap} and \textit{overlap} outcomes using the pre-turn window.
Both the gaze-only and multimodal variants were trained on identical LOGO/LOCO folds
(see Section \ref{Model Training and Evaluation}),
allowing paired comparison of fold-wise performance.
Across folds, the multimodal model achieved a mean ROC~AUC of \textbf{0.76~$\pm$~0.04} and PR~AUC of \textbf{0.77~$\pm$~0.05}, compared with \textbf{0.58~$\pm$~0.05} and \textbf{0.60~$\pm$~0.06} for the gaze-only model. 
The average paired improvements were $\Delta$AUC~=~\textbf{+0.18} (95\%\,CI\,[0.16,\,0.19]) and $\Delta$PR~=~\textbf{+0.17} ([0.14,\,0.19]), consistent across all folds (Wilcoxon~$p<0.001$). Figure \ref{fig:AUC_ModelPerf_Gaze_Phon}
 shows the ROC AUC scores between the two models.

These results confirm that adding speech-intensity information substantially improved predictive separability between gap and overlap outcomes.

\subsection{Added Value of Gaze and Phonetic Features}

\begin{figure}[h]
    \centering
    \includegraphics[width=\linewidth]{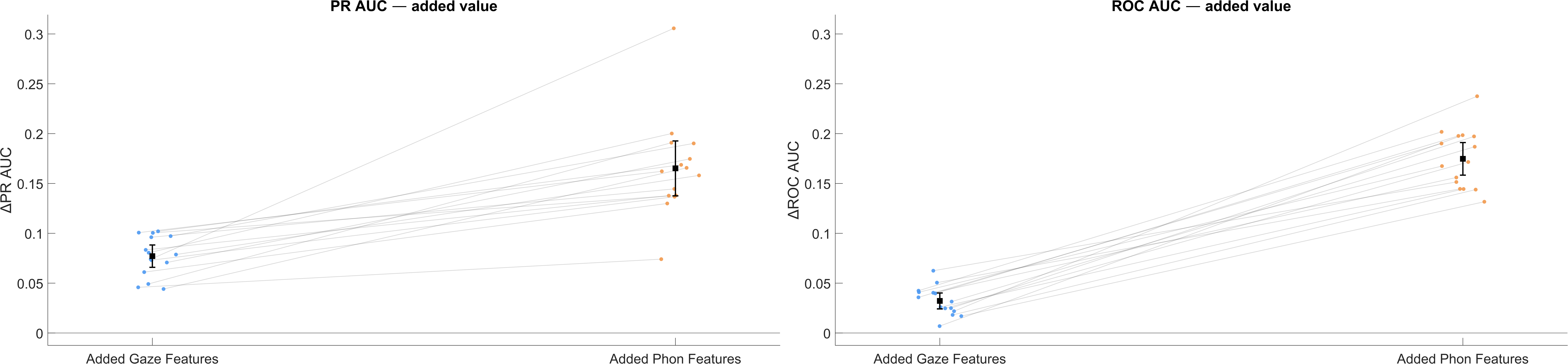}
    \caption{Added value of gaze and phonetic features in predicting floor-transfer outcomes. 
    Left: $\Delta$PR--AUC; Right: $\Delta$ROC--AUC. 
    Black squares show mean~$\pm$~95\%~CI across folds.}
    \label{fig:AddedValue_ROC_PR_Gaze_Phon}
\end{figure}

To quantify the contribution of each modality, we compared the multimodal model (PT) with ablations excluding gaze or phonetic features.
Figure~\ref{fig:AddedValue_ROC_PR_Gaze_Phon} shows paired fold-wise changes in PR--AUC and ROC--AUC, with each line linking the same cross-validation fold and black squares marking mean~$\pm$~95\%~CI.

\begin{figure}[h]
    \centering
    \includegraphics[width=1\linewidth, trim={0 0 0 0cm}, clip]{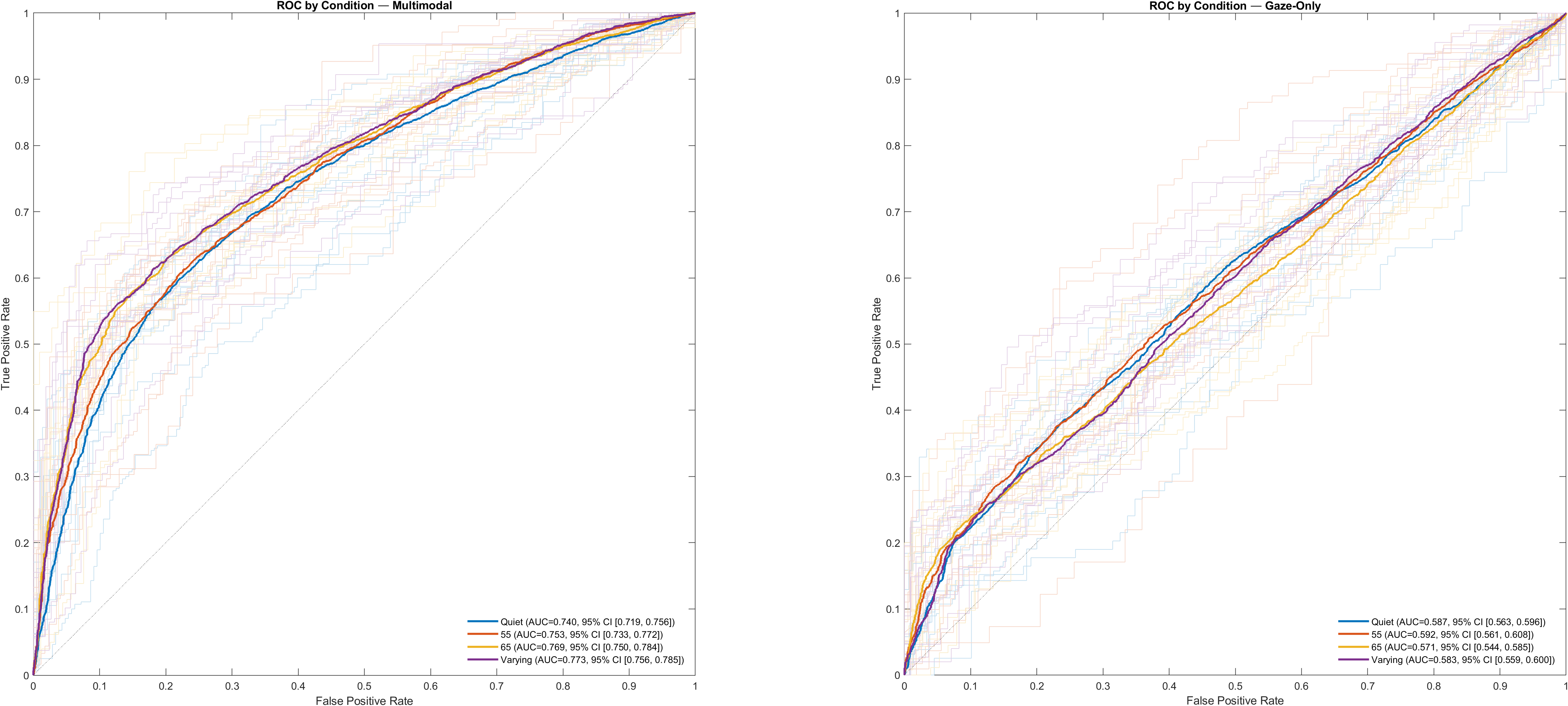}\\[4pt]
    \caption{Receiver operating characteristic (ROC) curves by acoustic condition for gaze-only and multimodal models. 
    Colors indicate acoustic conditions (Quiet, 55~dB, 65~dB, and~Varying). 
    Bold lines show pooled performance across outer folds, while faint "ghost" lines represent individual folds. 
    Multimodal models consistently achieved higher AUCs (0.74–0.77) than gaze-only (0.56–0.59), demonstrating robust generalisation across noise conditions.}
    \label{fig:ROC_Gap_Gaze_Multi}
\end{figure}

\begin{figure}[h]
    \centering
    \includegraphics[width=1\linewidth, trim={0 0 0 0cm}, clip]{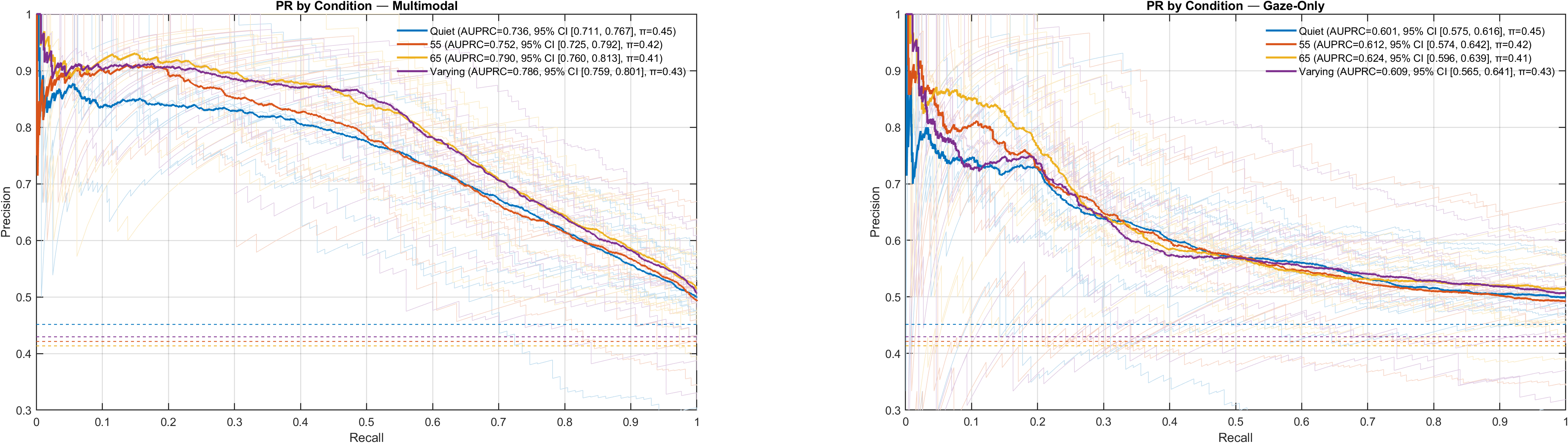}
  \caption{Precision–recall (PR) curves by acoustic condition for gaze-only and multimodal models. 
    Colors mark acoustic conditions; bold lines show the pooled performance and faint lines individual folds. 
    Dashed horizontal lines denote the class baseline precision. 
    Gaze features alone provided consistent predictive structure, while adding phon intensity yielded earlier recall gains and higher overall precision across all conditions.}
    \label{fig:PR_Gap_Gaze_Multi}
\end{figure}

\newpage
Adding gaze to the phon-only baseline improved PR-AUC by \textbf{+0.08$\pm$0.02} and ROC-AUC by \textbf{+0.03} (all folds positive, Wilcoxon $p<0.001$). 
Adding phon features to the gaze-only baseline produced larger gains (\textbf{+0.17} in both metrics). 
Thus, while phon intensity remained the dominant cue, gaze provided a consistent precision-oriented improvement, enhancing early recall and stabilizing predictions across acoustic conditions.
\\\\
Figures~\ref{fig:ROC_Gap_Gaze_Multi} and \ref{fig:PR_Gap_Gaze_Multi} summarise ROC and PR curves by acoustic condition. 
Multimodal models achieved AUCs of~0.74–0.77 across conditions, compared with~0.56–0.59 for gaze-only, demonstrating robust generalisation to noise.

\newpage
At the best-F1 threshold, the multimodal model reached \textbf{Precision = 0.62}, \textbf{Recall = 0.80}, \textbf{F1 = 0.69}, and \textbf{Accuracy = 0.71}, compared with 0.50, 1.00, 0.66, and 0.58 for gaze-only. 
Phon intensity improved balance and separability, while gaze remained robust across noise levels. 
Together, these results indicate that speech intensity signals speaker hold strength, whereas gaze reflects recipient readiness and occasional dominance cues preceding competitive entries.

\subsection{Performance Across Acoustic Conditions}

\begin{figure}[!h]
    \centering
    \includegraphics[width=1\linewidth]{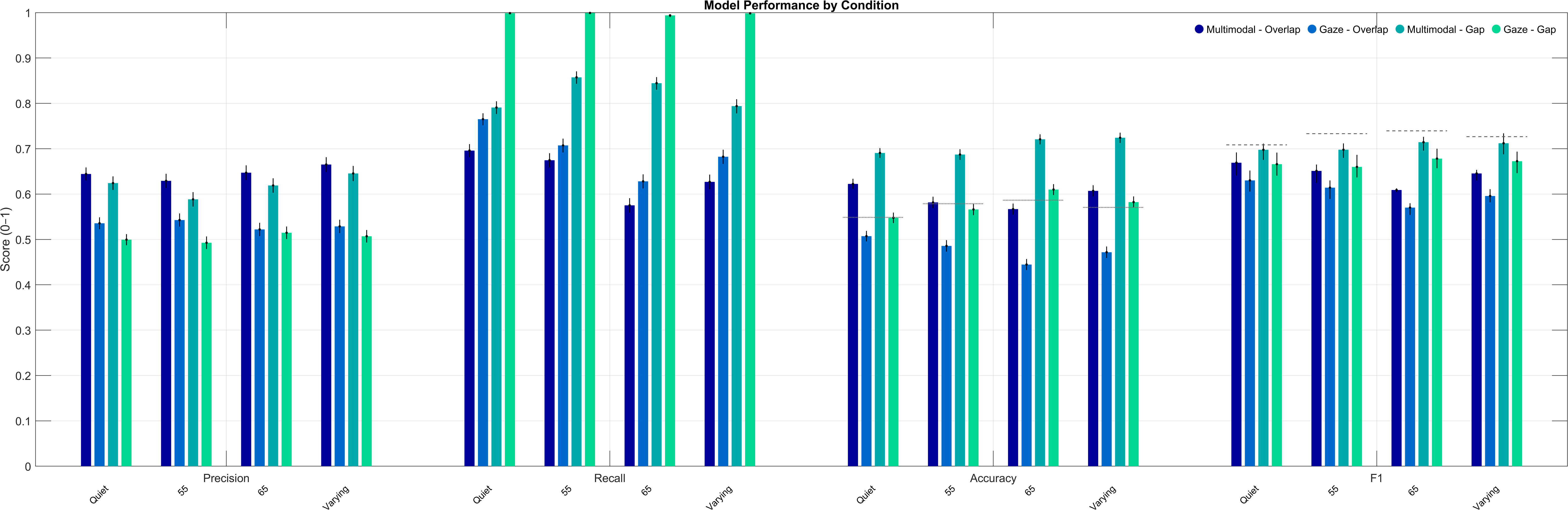}
    \caption{Per-condition performance metrics (Precision, Recall, Accuracy, and~F1) for gaze-only and multimodal models, separated by outcome type (gap or overlap). 
    Bars show mean scores across outer folds; error bars indicate standard error. Metrics are reported at the best-F1 threshold per fold.}
    \label{fig:ModelPerformance_Gaze_Multi}
\end{figure}

We next examined how model performance generalised across noise environments. 
Figure~\ref{fig:ModelPerformance_Gaze_Multi} summarises Precision, Recall, Accuracy, and~F1-scores for the gaze-only and multimodal models, separated by outcome type (gap or overlap) and acoustic condition (Quiet, 55~dB, 65~dB, Varying). Performance patterns were stable across all acoustic conditions. 
The multimodal model consistently achieved higher Precision and Accuracy (by~0.10–0.15) and slightly higher F1, while Recall remained high for both models. 

\newpage
Interestingly, accuracy and F1 tended to increase under the 65~dB and~\textit{Varying} conditions, indicating adaptive conversational adjustments rather than degradation by noise. 
Thus, the multimodal advantage observed in aggregated AUC and~AUPRC metrics (Section~\ref{Overall Predictive Performance}) persisted across environments, confirming that both gaze and speech intensity provide complementary, noise-robust cues to conversational state.

\subsection{Feature Contributions and Stability}

To identify which cues contributed most strongly to model predictions, we examined feature influence and stability across folds. 
Phon intensity windows (200--600 ms) formed the dominant feature block, confirming that short-term loudness provided the main signal for discriminating between gap and overlap outcomes. 
Across folds, these features yielded the largest odds ratios (ORs) and were retained in every model. 
In both model variants, gaze entropy metrics (SGE and GTE) were among the most frequently selected predictors, indicating that variability in gaze distribution and coordination captured meaningful information about conversational state. 
The addressing feature (\textit{Addressed - Weighted}) remained a robust non-verbal cue, reflecting recipient orientation and potential floor-transfer attempts.

\begin{figure}[!h]
    \centering
    \begin{subfigure}[!h]{0.4\linewidth}
        \centering
        \includegraphics[width=\linewidth]{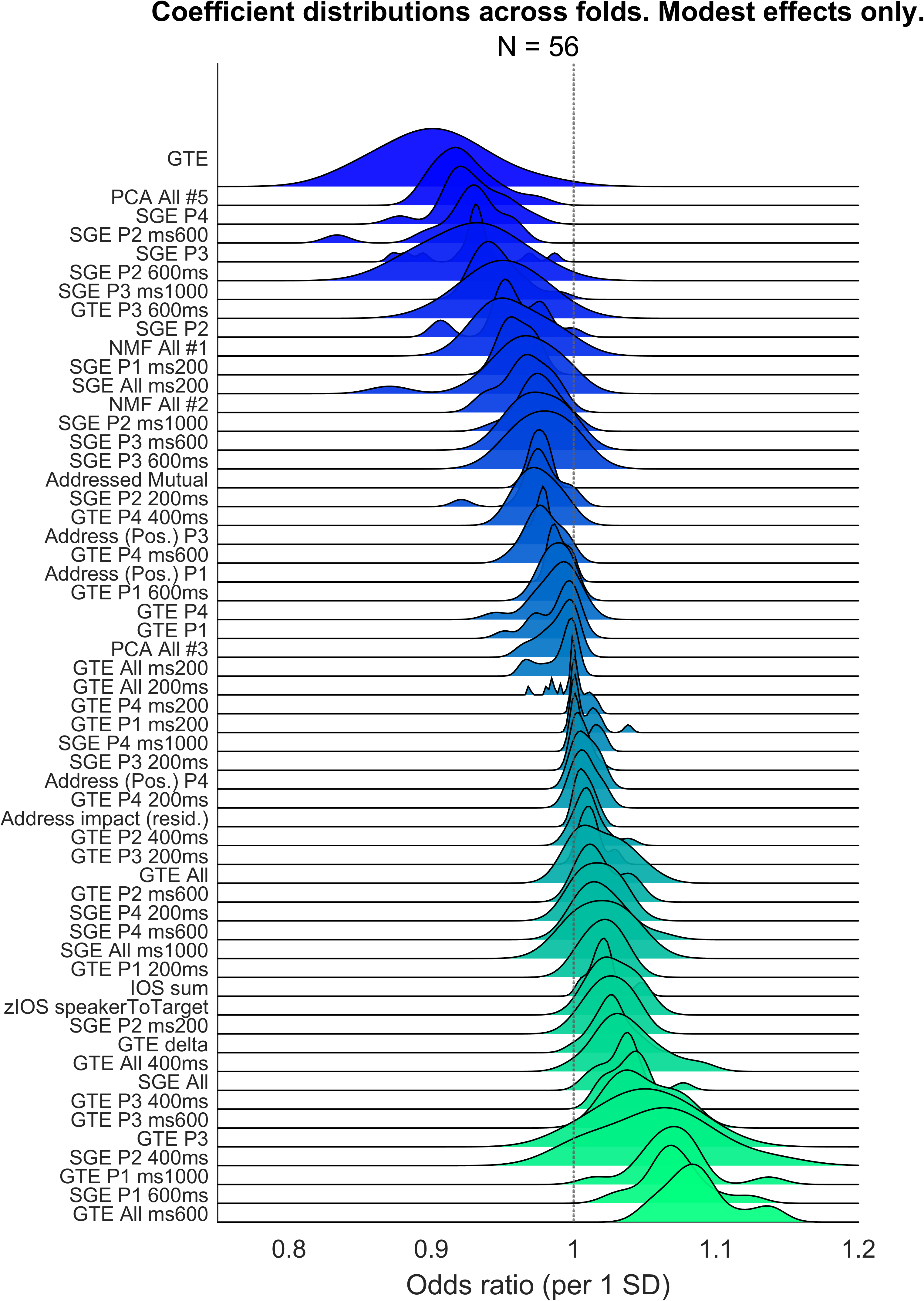}
        \caption{Gaze-only model}
        \label{fig:OR_modest_gaze}
    \end{subfigure}
    \hfill
    \begin{subfigure}[!h]{0.4\linewidth}
        \centering
        \includegraphics[width=\linewidth]{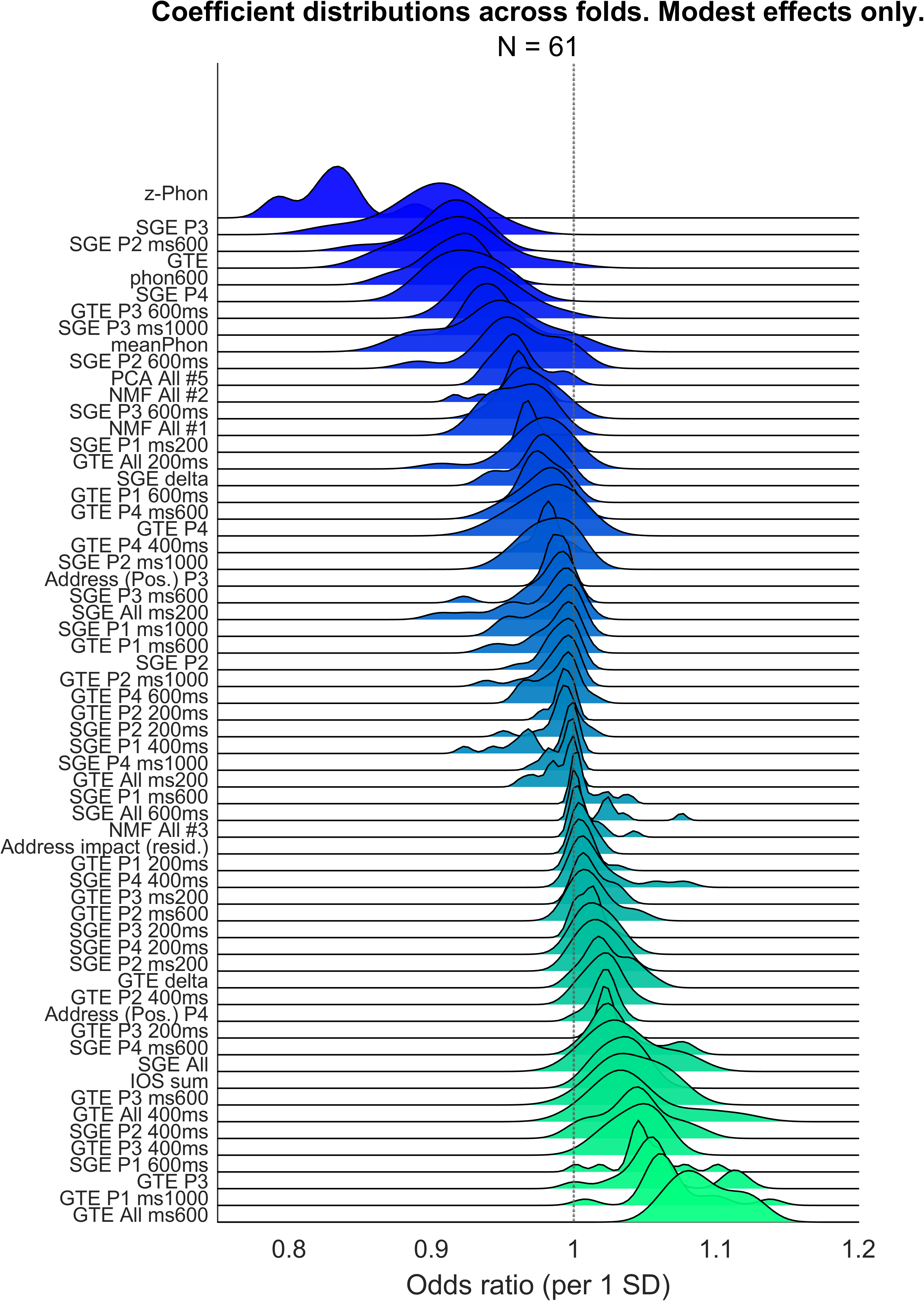}
        \caption{Multimodal model}
        \label{fig:OR_modest_multi}
    \end{subfigure}
    \caption{Fold-level odds ratios (per 1 SD) for stable predictors in the gaze-only (a) and multimodal (b) models.  
    Densities indicate the distribution of effects across folds; values above 1 favour \textit{overlap}, values below 1 favour \textit{gap}.}
    \label{fig:joyplots_sidebyside}
\end{figure}

Figures \ref{fig:OR_modest_gaze}–\ref{fig:OR_modest_multi} show the distribution of fold-level coefficients (expressed as OR per standard deviation) for a subset of stable features in the gaze-only and multimodal models. Volatile coefficients with high deviance were excluded.

The majority of features clustered near an OR of 1, with modest but consistent deviations in the direction of overlap prediction for SGE, GTE, and addressing cues.  
Although individual effects were small, their stability and additive influence across folds collectively support the cross-validated performance reported above.

To assess polarity dependence, we compared selection frequencies between overlap-positive and gap-positive fits of the gaze-only model.  
Features more often selected in the overlap-positive fits included entropy measures (SGE, GTE), the \textit{baseline} addressing featureset (i.e. no mutual gaze), and short-window variants, suggesting that dispersed and recipient-oriented gaze patterns are more diagnostic of overlap than of gap behaviour.
\subsection{Directionality of Strong Predictors}

\begin{figure}[h]
    \centering
    \includegraphics[width=1\linewidth]{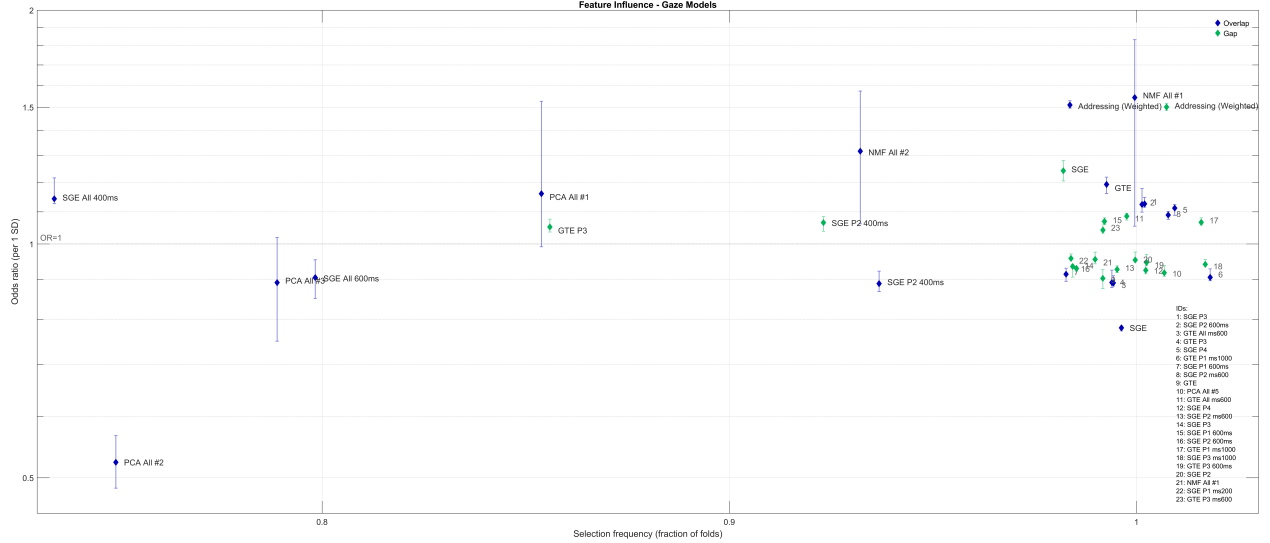}
    \caption{Feature influence for the gaze-only model shown jointly for \textit{overlap}-positive (blue) and \textit{gap}-positive (green) fits.
Each marker shows the median odds ratio (OR, per~1~SD) with interquartile range across folds, plotted against selection frequency.
Values above~1 indicate overlap bias; values below~1 indicate gap bias. Horizontal jitter has been added for visual clarity.}
    \label{fig:FeatureFreqOR_OverlapGap_Gaze}
\end{figure}

We next examined the direction and magnitude of feature effects in both the gaze-only and multimodal models.
Figure \ref{fig:FeatureFreqOR_OverlapGap_Gaze} shows the gap- and overlap-positive fits combined, for the gaze-only model, to visualise how feature influence differed with polarity.
Values above the horizontal line (OR~=~1) indicate greater likelihood of \textit{overlap}, and values below~1 indicate greater likelihood of \textit{gap}.

A clear separation emerged among the component-level features.
Non-negative matrix factorisation (NMF) components (\textit{NMF All \#1--\#3}) exhibited high selection frequency and elevated ORs toward \textit{overlap}, indicating that recurrent gaze-transition motifs are especially diagnostic of competitive entry.
Principal component axes (\textit{PCA All \#1--\#3}) showed mixed associations. Some were aligned with overlap, others with gap, which is consistent with their role as orthogonal variance contrasts.
Together, these findings suggest that turn competition and timing are encoded in both motif-like (NMF) and variance-based (PCA) structures:
NMF captures recurring, interpretable transition motifs, whereas PCA reflects broader variance in gaze allocation.

\begin{figure}[!ht]
    \centering
    \includegraphics[width=1\linewidth]{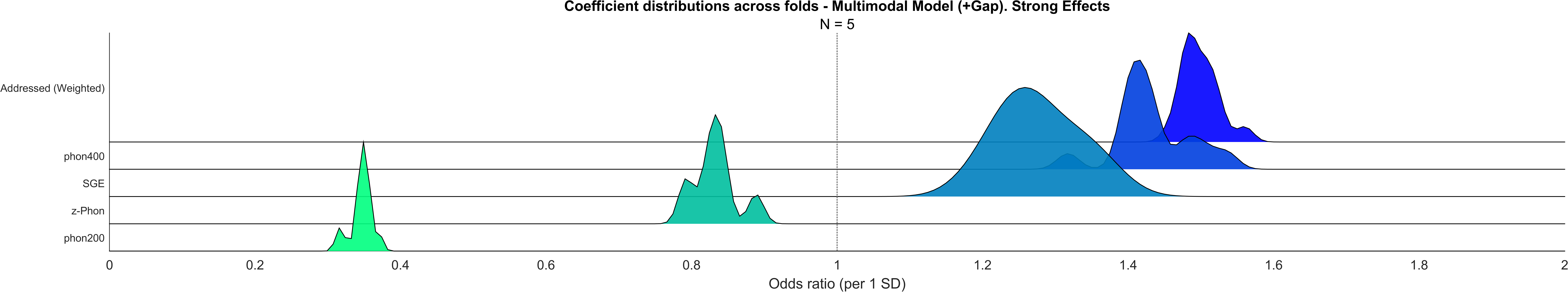}
    \caption{Direction and magnitude of strong predictors across modalities.
Bars show median odds ratios per~1~SD with interquartile ranges across folds. 
Phon intensity decreases overlap likelihood (floor-management), whereas gaze-related features increase it (transition relevance place).}
    \label{fig:joyPlot_strong}
\end{figure}

Baseline gaze cues supported this pattern.
Entropy measures (SGE, GTE) and addressing behaviour (\textit{Addressed - Weighted}) retained consistent directionality across folds, with higher values linked to increased overlap probability.
These correspond to dispersed or recipient-oriented gaze, characteristic of turn-competitive contexts.
The convergence of component-level and baseline effects strengthens the view that overlap events are preceded by structured yet flexible gaze dynamics, while gap events involve more stable visual behaviour. Figure \ref{fig:joyPlot_strong} shows the top five strongest features for gap as the positive outcome variable.

In the multimodal model, phon intensity remained the dominant predictor, with high values decreasing the probability of overlap and signalling continued speaker control.
However, gaze-derived predictors preserved the same directional tendencies observed in the gaze-only model, albeit with smaller relative magnitude.

This consistency indicates that gaze and speech convey complementary information:
vocal intensity encodes speaker hold strength, while gaze patterns capture recipient design and readiness to enter the conversational floor.

\section{Discussion}

This study investigated the predictive value of multimodal cues for conversational floor-transfer outcomes in natural four-person dialogue. 
By integrating gaze and speech features with interpersonal closeness, within a logistic framework, we demonstrated that non-verbal visual behaviour contributes consistently to the discrimination of \textit{gap} and \textit{overlap} events.
Four contributions emerge from the results.
First, the models provide a predictive account of multimodal behaviour, showing that gaze-derived features alone can forecast conversational outcomes.
Second, multimodal integration yielded reliable improvements across speakers and acoustic conditions, confirming the robustness of gaze cues even in noise.
Third, the analysis isolates gaze features, including entropy, addressing, and component-level structure, as stable, precision-oriented predictors of turn dynamics.
Finally, the use of component decomposition (PCA and NMF) introduced an interpretable representation of recurrent gaze transitions, extending prior descriptive work on visual coordination into a predictive domain.

\paragraph{Interpretation of multimodal advantage.}
Across folds and conditions, the multimodal model consistently outperformed the gaze-only baseline, driven primarily by phon intensity yet complemented by gaze-based cues. 
Phon intensity captured the speaker’s control or persistence in holding the conversational floor, while gaze captured anticipatory and recipient-oriented behaviour.
This division of labour aligns with theoretical models of turn organisation, in which participants continually negotiate entry and exit through multimodal signals (Levinson \& Torreira,~2015).
Gaze shifts reflected both recipient readiness and pre-emptive engagement during turn competition, complementing the acoustic indication of speaker control.
Together, these modalities may capture aspects of the balance between speaker continuation and listener approach to the floor. 

\paragraph{Interpersonal Context}
In addition to moment-to-moment gaze and speech behaviour, perceived interpersonal closeness provided a relational description of the participants involved in each floor transfer. Unlike the temporally varying behavioural features, IOS was static within a dyad and directional, allowing the model to account for differences in the social relationship between outgoing and incoming speakers. While the present analysis does not examine how interpersonal closeness interacts or covaries with specific gaze and speech behaviours, its inclusion provides a social-contextual component alongside the behavioural predictors. Understanding whether relational structure is expressed through particular gaze motifs, addressing patterns, or other turn-taking behaviours remains an important direction for subsequent analysis.

\paragraph{Component-level structure in gaze dynamics.}
Beyond individual gaze features, the component analysis revealed that overlap and gap events are distinguished by their underlying visual transition patterns.
Principal component axes behaved as broad variance contrasts, while non-negative matrix factorisation (NMF) components captured motif-like, additive structures that were strongly associated with overlap.
Importantly, these components were derived from symbolic gaze \emph{n}-grams computed directly from continuous gaze trajectories rather than from annotated tiers, allowing unsupervised discovery of recurrent visual motifs.
This result suggests that overlap events are preceded by structured, recurrent gaze sequences, whereas gap events are associated with more stable, homogeneous gaze behaviour.
The combination of PCA and NMF thus offers a compact yet interpretable representation of gaze coordination at the transition point.

We also observed that the addressing cue showed a positive association with \textit{gap} in the gap-positive analyses. One plausible interpretation is that explicit addressee orientation can create a form of “forced candidacy”: the selected recipient needs a short preparation interval for utterance planning and local TRP estimation, yielding a brief pause before uptake. This is consistent with the idea that recipient selection can introduce planning latency on the order of a few hundred milliseconds producing non-competitive timing rather than immediate overlap. We view this mechanism as a data-specific explanation rather than a universal account, since its expression likely depends on condition and local interactional context.

\paragraph{Speech feature-space and methodological integrity.}
The speech feature-space was deliberately sparse to avoid tautological learning from diarisation or overlap information inherently present in the corpus.
Phon intensity features were computed only for the current speaker and were temporally aligned to the speaker’s offset, excluding any candidate or future-speaker information.
This design minimised the risk of information leakage but necessarily limited the raw AUC values for speech-based models.
Nevertheless, the modest but consistent discriminability of the gaze models indicates that gaze features capture meaningful structure even without access to direct acoustic proxies for turn boundaries. 
Earlier control experiments confirmed that when candidate-speaker or offset-aligned phon features were included, the models reached unrealistically high AUROCs ($>$0.90), illustrating how tightly controlled features are essential for conceptual validity.
All reported models were trained and evaluated on held-out groups and conditions (LOGO/LOCO), ensuring out-of-sample generalisability.

\paragraph{Robustness, limitations, and outlook.}
Performance patterns were consistent across all acoustic conditions, indicating that gaze-based cues remain reliable even under degraded speech environments. 
The smaller variance observed for condition-wise evaluation may partly reflect data availability: the LOCO splits divide the dataset by one of four acoustic conditions, whereas the LOGO splits partition by one of eleven groups, leaving less training data per fold in the latter case. 
Groups may also exhibit idiosyncratic, aggregated behaviour that disappears when a single group is held out, while LOCO splits always include contributions from all groups. 
This difference in sampling composition likely contributes to the greater variability across LOGO folds and to the occasional over- or under-performing folds visible as “ghost lines’’ in the ROC curves.

The study was conducted in a controlled conversational setting but employed purposive sampling of native (L1) speakers engaged in unscripted, free conversation without imposed tasks or turn-taking constraints. 
This design ensured natural interactional dynamics while maintaining control over recording conditions and participant homogeneity. 
The models generalised across speaker groups, supporting the robustness of the feature representations despite limited data. 
Nevertheless, several limitations remain. 
Sampling constraints in the gaze and audio streams limited temporal resolution, and the feature-engineered windows may underrepresent the dynamics occurring outside the pre-turn interval. 
Future work should extend predictive modelling beyond static windows toward temporally continuous, context-aware approaches that integrate additional non-verbal cues such as head orientation or gestural motion. 
Incorporating conversational structure and linguistic context could further clarify how visual and acoustic behaviours jointly forecast floor-transfer outcomes in real-world interaction. 
More broadly, predictive multimodal modelling of turn-taking holds promise for hybrid and agent-mediated communication systems that aim to facilitate smoother coordination between human interlocutors and artificial agents.

\section{Conclusion}
This study presented a predictive, multimodal model of conversational floor-transfer outcomes, combining gaze and speech cues with perceived interpersonal closeness in natural four-person dialogue.
Gaze features, including entropy, addressing, and component-level transitions, provided consistent, precision-oriented information that complemented speech intensity in predicting whether turn boundaries would result in a gap or overlap. 
Across speaker groups and acoustic conditions, the multimodal model achieved robust generalisation while deliberately avoiding feature leakage from diarisation or candidate speakers. 
These results indicate that non-verbal gaze behaviour carries predictive structure relevant to turn-taking organisation and contributes to the coordination of conversational control. 
Unlike speech cues, which depend on signal-to-noise ratio and occur in discrete units, gaze provides a continuous behavioural stream that remains observable even when acoustic information is degraded. 
This distinction underscores the complementary role of gaze in understanding conversational dynamics and highlights its potential value for future multimodal modelling of natural interaction.

\section*{Data Availability}


The data analysed in this study are publicly available as part of the GaMMA corpus via Zenodo \url{https://doi.org/10.5281/zenodo.18343509}\cite{DouradoGaMMA_Dryad}. A detailed description of the dataset, including data collection procedures, processing pipelines, ethics approval, and usage notes, is provided in the associated data descriptor publication \cite{DouradoGaMMA2025}. 

\section*{Code Availability}
The analysis code used in this study was developed as part of an industrial research collaboration and is subject to intellectual property restrictions. As such, the full codebase cannot be made publicly available at this time. The analysis procedures and modelling steps are described in sufficient detail in the Methods section to allow independent replication.

\section*{Competing Interests}
The authors declare no competing interests.

\section*{Funding}
This research was supported by Innovation Fund Denmark through an industrial research collaboration with GN Store Nord.

\section*{Author Contributions}
Mark Dourado conceived the study, performed the analysis, and wrote the manuscript. All authors reviewed and approved the final version of the manuscript.

\bibliographystyle{plain}
\bibliography{JP2Bibtex}

@article{goodwin1981conversational,
  title={Conversational organization},
  author={Goodwin, Charles},
  journal={Interaction between speakers and hearers},
  year={1981},
  publisher={Academic press}
}

@article{aron1992IOS,
  title={Inclusion of other in the self scale and the structure of interpersonal closeness.},
  author={Aron, Arthur and Aron, Elaine N and Smollan, Danny},
  journal={Journal of personality and social psychology},
  volume={63},
  number={4},
  pages={596},
  year={1992},
  publisher={American Psychological Association}
}

@misc{zwicker:iso,
  author       = {{International Organization for Standardization}},
  year         = {2017},
  title        = {{ISO 532-1:2017 — Acoustics — Methods for calculating loudness — Part 1: Zwicker method}},
  address      = {Geneva, Switzerland},
  howpublished = {International Organization for Standardization},
  note         = {Retrieved from \url{https://cdn.standards.iteh.ai/samples/63077/58c7f464f167413abde1367b3e0906db/ISO-532-1-2017.pdf}}
}

@article{sorensen2021effects,
  title={Effects of noise and second language on conversational dynamics in task dialogue},
  author={S{\o}rensen, A Josefine Munch and Fereczkowski, Michal and MacDonald, Ewen N},
  journal={Trends in Hearing},
  volume={25},
  pages={23312165211024482},
  year={2021},
  publisher={SAGE Publications Sage CA: Los Angeles, CA},
  note = {DOI: https://doi.org/10.1177/23312165211024482}
}

@article{heldner2010pauses,
  title={Pauses, gaps and overlaps in conversations},
  author={Heldner, Mattias and Edlund, Jens},
  journal={Journal of Phonetics},
  volume={38},
  number={4},
  pages={555--568},
  year={2010},
  publisher={Elsevier}
}

@article{ford1996,
  title={Practices in the construction of turns: The “TCU” revisited},
  author={Ford, Cecilia E and Fox, Barbara A and Thompson, Sandra A},
  journal={Pragmatics. Quarterly Publication of the International Pragmatics Association (IPrA)},
  volume={6},
  number={3},
  pages={427--454},
  year={1996},
  publisher={John Benjamins Publishing Company Amsterdam/Philadephia}
}

@article{ho2015speaking,
  title={Speaking and listening with the eyes: Gaze signaling during dyadic interactions},
  author={Ho, Simon and Foulsham, Tom and Kingstone, Alan},
  journal={PloS one},
  volume={10},
  number={8},
  pages={e0136905},
  year={2015},
  publisher={Public Library of Science San Francisco, CA USA}
}

@article{slomianka2025adaptions,
  title={Adaptions in eye-movement behavior during face-to-face communication in noise},
  author={Slomianka, Valeska and May, Tobias and Dau, Torsten},
  journal={Frontiers in Psychology},
  volume={16},
  pages={1584937},
  year={2025},
  publisher={Frontiers}
}

@article{stevens1955loudness,
  title={The measurement of loudness},
  author={Stevens, Stanley S},
  journal={The Journal of the Acoustical Society of America},
  volume={27},
  number={5},
  pages={815--829},
  year={1955},
  publisher={Acoustical Society of America}
}

@article{duncan1972some,
  title={Some signals and rules for taking speaking turns in conversations.},
  author={Duncan, Starkey},
  journal={Journal of personality and social psychology},
  volume={23},
  number={2},
  pages={283},
  year={1972},
  publisher={American Psychological Association}
}

@article{clark2002using,
  title={Using uh and um in spontaneous speaking},
  author={Clark, Herbert H and Tree, Jean E Fox},
  journal={Cognition},
  volume={84},
  number={1},
  pages={73--111},
  year={2002},
  publisher={Elsevier}
}

@article{gravano2011turn,
  title={Turn-taking cues in task-oriented dialogue},
  author={Gravano, Agust{\'\i}n and Hirschberg, Julia},
  journal={Computer Speech \& Language},
  volume={25},
  number={3},
  pages={601--634},
  year={2011},
  publisher={Elsevier}
}

@article{caspers2003,
  title={Local speech melody as a limiting factor in the turn-taking system in Dutch},
  author={Caspers, Johanneke},
  journal={Journal of Phonetics},
  volume={31},
  number={2},
  pages={251--276},
  year={2003},
  publisher={Elsevier}
}

@article{Hadley2019,
   author = {Lauren V. Hadley and W. Owen Brimijoin and William M. Whitmer},
   doi = {10.1038/s41598-019-46416-0},
   issn = {20452322},
   issue = {1},
   journal = {Scientific Reports},
   month = {12},
   pmid = {31320658},
   publisher = {Nature Publishing Group},
   title = {Speech, movement, and gaze behaviours during dyadic conversation in noise},
   volume = {9},
   year = {2019},
}

@article{Garnier2010,
   author = {Maëva Garnier and Nathalie Henrich Bernardoni and Danièle Dubois},
   doi = {10.1044/1092},
   issue = {3},
   journal = {American Speech-Language-Hearing Association},
   pages = {588-608},
   title = {Influence of Sound Immersion and Communicative Interaction on the Lombard Effect},
   volume = {53},
   url = {https://hal.archives-ouvertes.fr/hal-00480027},
   year = {2010},
}

@article{Trujillo2021,
   author = {James Trujillo and Asli Özyürek and Judith Holler and Linda Drijvers},
   doi = {10.1038/s41598-021-95791-0},
   issn = {20452322},
   issue = {1},
   journal = {Scientific Reports},
   month = {12},
   pmid = {34408178},
   publisher = {Nature Research},
   title = {Speakers exhibit a multimodal Lombard effect in noise},
   volume = {11},
   year = {2021},
}

@article{Ishii2016,
   author = {Ryo Ishii and Kazuhiro Otsuka and Shiro Kumano and Junji Yamato},
   doi = {10.1145/2757284},
   issn = {21606463},
   issue = {1},
   journal = {ACM Transactions on Interactive Intelligent Systems},
   month = {5},
   publisher = {Association for Computing Machinery},
   title = {Prediction of who will be the next speaker and when using gaze behavior in multiparty meetings},
   volume = {6},
   year = {2016},
}

@article{Stivers2009,
  title={Universals and cultural variation in turn-taking in conversation},
  author={Stivers, Tanya and Enfield, Nicholas J and Brown, Penelope and Englert, Christina and Hayashi, Makoto and Heinemann, Trine and Hoymann, Gertie and Rossano, Federico and De Ruiter, Jan Peter and Yoon, Kyung-Eun and others},
  journal={Proceedings of the National Academy of Sciences},
  volume={106},
  number={26},
  pages={10587--10592},
  year={2009},
  publisher={National Academy of Sciences}
}

@misc{Sacks1974,
   author = {Harvey Sacks and Emanuel A Schegloff and Gail Jefferson},
   isbn = {130.225.53.20},
   issue = {4},
   journal = {Source: Language},
   pages = {696-735},
   title = {A Simplest Systematics for the Organization of Turn-Taking for Conversation},
   volume = {50},
   year = {1974},
}

@inproceedings{Ishii2013,
   author = {Ryo Ishii and Kazuhiro Otsuka and Shiro Kumano and Masafumi Matsuda and Junji Yamato},
   doi = {10.1145/2522848.2522856},
   isbn = {9781450321297},
   booktitle = {ICMI 2013 - Proceedings of the 2013 ACM International Conference on Multimodal Interaction},
   pages = {79-86},
   title = {Predicting next speaker and timing from gaze transition patterns in multi-party meetings},
   year = {2013},
}

@article{shiferaw2019gaze,
  author       = {Shiferaw, Behailu and Crewther, David P. and Downey, Lachlan A.},
  title        = {A review of gaze entropy as a measure of visual scanning efficiency},
  journal      = {Neuroscience \& Biobehavioral Reviews},
  volume       = {98},
  year         = {2019},
  pages        = {218--237},
  doi          = {10.1016/j.neubiorev.2018.12.007},
}

@inproceedings{krejtz2015entropy,
  author       = {Krejtz, Krzysztof and Duchowski, Andrew and Szmidt, Tomasz and Krejtz, Izabela and González Perilli, Fernando and Pires, Ana and Vilaro, Anna and Villalobos, Natalia},
  title        = {Gaze Transition Entropy: Gaze Transitions as Markov Chains and Shannon's Entropy Coefficient},
  booktitle    = {Proceedings of the ACM Symposium on Eye Tracking Research \& Applications (ETRA)},
  year         = {2015},
  pages        = {187--190},
  doi          = {10.1145/2735713.2735744},
}

@article{LevinsonTorreira2015,
  author    = {Stephen C. Levinson and Francisco Torreira},
  title     = {Timing in turn-taking and its implications for processing models of language},
  journal   = {Frontiers in Psychology},
  year      = {2015},
  volume    = {6},
  pages     = {731}
}

@article{HollerKendrick2015,
  author    = {Judith Holler and Kobin H. Kendrick},
  title     = {Unaddressed participants’ gaze in multi-person interaction: Optimizing recipiency},
  journal   = {Frontiers in Psychology},
  year      = {2015},
  volume    = {6},
  pages     = {98}
}

@article{Summers1988,
  author    = {William V. Summers and David B. Pisoni and Robert H. Bernacki and Robert I. Pedlow and Michael A. Stokes},
  title     = {Effects of noise on speech production: Acoustic and perceptual analyses},
  journal   = {Journal of the Acoustical Society of America},
  year      = {1988},
  volume    = {84},
  number    = {3},
  pages     = {917--928}
}

@article{DouradoGaMMA2025,
  title={The GaMMA corpus of Danish polyadic conversations with gaze speech and motion data in quiet and noise},
  author={Dourado, Mark and Gert Hassager, Henrik and Udesen, Jesper and Serafin, Stefania},
  journal={Scientific Data},
  volume={13},
  number={1},
  pages={494},
  year={2026},
  publisher={Nature Publishing Group UK London},
  url ={https://doi.org/10.1038/s41597-026-06851-x},
  note = {DOI: https://doi.org/10.1038/s41597-026-06851-x}
}

@misc{DouradoGaMMA_Dryad,
  author       = {Mark Dourado and Henrik Gert Hassager and Jesper Udesen and Stefania Serafin},
  title        = {GaMMA Corpus: Polyadic Conversations in Danish with Gaze, Speech, and Motion Data in Quiet and Adverse Speech Conditions},
  year         = {2026},
DOI = {https://doi.org/10.5281/zenodo.18343509},
note = {Dataset. DOI: https://doi.org/10.5281/zenodo.18343509}
}

@software{Sorensen2025,
  author       = {A. Josefine Munch S{\o}rensen},
  title        = {Communicative State Classification: MATLAB algorithm for classifying multi-talker audio into timings and durations of conversational dynamics},
  year         = {2025},
  version      = {1.0.0},
  publisher    = {Zenodo},
  doi          = {10.5281/zenodo.17063065},
  url          = {https://doi.org/10.5281/zenodo.17063065},
  note         = {Computer software. DOI: https://doi.org/10.5281/zenodo.17063065}
}

\end{document}